%% file: 0_main.tex
\documentclass{article}
\usepackage[preprint,nonatbib]{neurips_2026}

\usepackage[utf8]{inputenc}
\usepackage[T1]{fontenc}
\usepackage{hyperref}
\usepackage[backend=biber,style=numeric,maxcitenames=2,maxbibnames=99,natbib=true]{biblatex}
\usepackage{url}
\usepackage{booktabs}
\usepackage{amsfonts}
\usepackage{amsmath}
\usepackage{amssymb}
\usepackage{makecell}
\usepackage{nicefrac}
\usepackage{microtype}
\usepackage{xcolor}
\usepackage{enumitem}
\usepackage{multirow}
\usepackage{tabularx}
\usepackage{array}
\usepackage{listings}
\usepackage{courier}
\usepackage{graphicx}
\usepackage{algorithm}
\usepackage{algpseudocode}
\usepackage{wrapfig}
\usepackage{xcolor}
\usepackage{pifont}
\usepackage{subcaption}

\newcommand{\cmark}{\textcolor{green!60!black}{\ding{51}}}
\newcommand{\xmark}{\textcolor{red!75!black}{\ding{55}}}

\usepackage{amsmath}
\usepackage{amssymb}

\usepackage{fvextra}
\usepackage[most]{tcolorbox}

\DefineVerbatimEnvironment{PromptBlock}{Verbatim}{
  fontsize=\footnotesize,
  breaklines=true,
  breakanywhere=true,
  breaksymbolleft={},
  breaksymbolright={},
  baselinestretch=0.95
}

\newtcolorbox{promptbox}[1]{
  enhanced,
  breakable,
  colback=white,
  colframe=gray!35,
  colbacktitle=gray!45,
  coltitle=white,
  title={#1},
  fonttitle=\bfseries\small,
  boxrule=0.8pt,
  arc=4pt,
  outer arc=4pt,
  left=8pt,
  right=8pt,
  top=6pt,
  bottom=6pt,
  toptitle=3pt,
  bottomtitle=3pt,
  lefttitle=8pt,
  righttitle=8pt,
  before skip=0.6em,
  after skip=0.8em
}

\newcommand{\method}{\textsc{PMD}}
\newcommand{\sdpo}{\textsc{SDPO}}
\newcommand{\opcd}{\textsc{OPCD}}
\newcommand{\oel}{\textsc{OEL}}

\title{Procedural Memory Distillation: Online Reflection for Self-Improving Language Models}

\author{%
  Ye Liu, Srijan Bansal, Bo Pang, Yang Li, \\ \textbf{Zeyu Leo Liu, Yifei Ming, Zixuan Ke, Shafiq Joty, Semih Yavuz} \\
  Salesforce AI Research \\
  \texttt{yeliu, srijanbansal, sjoty, syavuz @salesforce.com}
}

\begin{document}
\maketitle

\input{1_introduction_sdpo_memory}

\input{related_work}

\input{2_method_v3}

\input{3_experiment}

\section{Conclusion}
\label{sec:conclusion}
We presented \method{}, a procedural memory distillation framework that turns repeated training-time experience into reusable supervision for self-improving language models. \method{} moves beyond episode-local hindsight by building memory online from the learner's own trajectories, abstracting it into experience, insight, and behavior memory, and using this evolving memory to condition an on-policy self-teacher. Because memory is used only during training, the final student does not depend on external retrieval at inference time; recurring procedural knowledge is instead gradually internalized into the policy. Our experiments on \textsc{SciKnowEval} and \textsc{LiveCodeBench} show that this co-evolution of memory and policy improves over SDPO and over controls with frozen policies or fixed memories. These results suggest that self-improving models benefit not only from immediate feedback, but also from structured memory that records what they have tried, reflects on what worked or failed, and distills reusable behaviors into the model itself. 

\textbf{Limitations.} Our current evaluation focuses on repeated training over fixed task distributions, where problem-local experience can be accumulated across epochs. Although we evaluate two different verifiable domains, science reasoning and code generation, this setting is still narrower than long-horizon agentic environments in which memory, tools, and skills must evolve across heterogeneous tasks \citep{wang2023voyager,cheng2026mem2evolve,wu2025evolver}. We therefore do not claim that PMD fully solves online agent self-improvement across domains. Rather, our results isolate a more specific question: whether procedural knowledge extracted from repeated attempts can co-evolve with the policy and be distilled into memory-free inference behavior. Extending PMD from problem-local memory to broader behavior retrieval across tasks, and validating it on long-horizon agent benchmarks, are important directions for future work.

\printbibliography

\appendix
\input{4_appendix_v3}

\newpage
\end{document}

%% file: 1_introduction_sdpo_memory.tex
\begin{abstract}

Reinforcement learning with verifiable rewards (RLVR), along with recent self-distillation variants such as SDPO, evaluates each rollout against a verifier and updates the policy from that episode-level signal. However, the richer procedural information in the rollout is rarely retained or reused. Across episodes and epochs, the model repeatedly encounters related problems under a changing policy, producing cross-episode signals that episode-local updates cannot capture: which strategies consistently pass verification, which failure modes persist, which patterns recur. We propose \textbf{Procedural Memory Distillation (\method)}, which converts these cross-episode signals into \textit{reusable procedural memory} and distills it into the policy's weights during training. This memory functions as a training scaffold, absorbed into the policy itself, yielding a memory-free model at inference. {\method} organizes the memory at three levels of abstraction: raw trajectories, self-reflected strategies and lessons, and higher-level behavioral patterns that recur across problems, all extracted online from the model's own trajectories. A memory-conditioned self-teacher draws on the accumulated experience to supervise the student on its own rollouts, enabling student to progressively internalize procedural knowledge within its parameters. The central design principle is co-evolution: the policy generates rollouts that update the memory, and memory shapes the supervision that updates the policy. Empirically, across Qwen3-8B and OLMo3-Instruct-7B, \method{} improves over SDPO by $3.8$--$5.5\%$ on \textsc{SciKnowEval} and $7.9$--$13.6\%$ on \textsc{LiveCodeBench}. Co-evolution powers these gains: freezing either the memory or the policy trails {\method} by more than 10\% across \textsc{SciKnowEval} domains.

\end{abstract}

\section{Introduction}
\label{sec:intro}

The prevailing paradigm for preference optimization and reinforcement learning with verifiable rewards (RLVR) operates at the level of individual episodes; methods such as PPO~\citep{schulman2017ppo}, DPO~\citep{rafailov2023dpo} and GRPO~\citep{shao2024deepseekmath,guo2025deepseekr1} convert per-rollout preferences or outcome-based checks into learning signals. Each rollout receives a reward, feedback or hindsight correction; the policy gets updated accordingly; and the experience is discarded. This design is natural when episodes are independent. However, in practice, models repeatedly encounter the same or related problems across epochs under a continually evolving policy. These repeated interactions carry cross-episode signals that isolated, one-step updates cannot capture: which strategies pass verification, which failure modes persist, which patterns recur.

Recent work suggests that the learning signal available during training can be richer than a scalar reward alone. For instance, in Self-Distillation Policy Optimization (SDPO)~\citep{hubotter2026sdpo}, the current policy can act as self-teacher when conditioned on training-time context: textual feedback when available, or a successful sibling rollout from the same group when it is not.
This reflects a broader shift from offline distillation to on-policy distillation, where the student is trained on states it actually visits rather than static teacher demonstrations~\citep{hinton2015distilling,ross2011dagger,agarwal2024opd,gu2024minillm,song2026surveyopd}. 

While SDPO and related on-policy distillation approaches~\cite{agarwal2024opd,zhao2026opsd} help alleviate the sparse reward limitations of standard RLVR and the distributional mismatch of offline distillation, their updates remain episode-local: they do not systematically preserve what the model has discovered across earlier attempts. 
We propose \textbf{Procedural Memory Distillation (PMD)}, which converts these repeated attempts into reusable procedural memory and distills it into the policy's weights during training. {\method} builds the memory online from the model’s own rollouts, organized into three levels: experience (raw trajectories), insight (per-problem strategies and lessons), and behavior (cross-problem reasoning patterns). 
This memory conditions an on-policy self-teacher: the student continues to learn from trajectories sampled from the current policy, as in SDPO, but the teacher is additionally informed by what the model has discovered in prior epochs.
Memory thus functions as a training scaffold: it enriches the teacher during learning and is gradually internalized into the student's parameters, so the resulting model reasons natively at inference.

A central design principle of PMD is the \emph{co-evolution of policy and memory}. 
At each stage of training, the current policy generates rollouts, receives feedback, and writes new experience into memory, and the updated memory then conditions the self-teacher that trains the next version of the policy. 
As shown on Table \ref{tab:conceptual_comparison}, this tight, online coupling is what distinguishes PMD from static or offline memory banks extracted once by a fixed model: as the learner evolves, memory evolves with it, keeping the teacher signal aligned with the policy's current strengths and failure modes throughout training.

PMD organizes procedural memory into three levels. \emph{Experience memory} (Level-0) stores rollouts, rewards, and feedback for each problem. \emph{Insight memory} (Level-1) reflects on these attempts to extract strategies and lessons. \emph{Behavior memory} (Level-2) groups semantically related problems and distills their experiences and insights into reusable reasoning patterns. This hierarchy exposes a trade-off between concreteness and abstraction. Experience memory preserves the faithful evidence but remains highly local, while behavior memory transfers broadly but can become too coarse. Insight memory strikes a middle ground by retaining problem-grounded lessons in a compact and reusable form. During training, PMD draws on problem-specific experience and insight memories, while retrieving cross-problem behavioral patterns from a global memory.

\begin{wraptable}{r}{0.58\linewidth}
\vspace{-1.2em}
\centering
\scriptsize
\setlength{\tabcolsep}{2.0pt}
\renewcommand{\arraystretch}{1.05}
\caption{
Comparison of learning and memory paradigms. Inference-time memory agent methods include
MemoryBank, MemGPT, Memento and A-MEM
\citep{zhong2024memorybank,packer2023memgpt,zhou2025memento,xu2026mem}.
}
\label{tab:conceptual_comparison}
\resizebox{\linewidth}{!}{
\begin{tabular}{@{}lccccc@{}}
\toprule
\textbf{Method}
& \makecell{\textbf{Evolving}\\\textbf{policy}}
& \makecell{\textbf{Memory-free}\\\textbf{inference}}
& \makecell{\textbf{Persistent}\\\textbf{memory}}
& \makecell{\textbf{Evolving}\\\textbf{memory}}
& \makecell{\textbf{Policy-memory}\\\textbf{co-evolution}} \\
\midrule
Base model
& \xmark & \cmark & \xmark & \xmark & \xmark \\
GRPO / RLVR~\citep{shao2024deepseekmath,guo2025deepseekr1}
& \cmark & \cmark & \xmark & \xmark & \xmark \\
SDPO / OPD~\citep{hubotter2026sdpo,zhao2026opsd,sang2026opsdc}
& \cmark & \cmark & \xmark & \xmark & \xmark \\
OPCD~\citep{ye2026opcd}
& \cmark & \cmark & \xmark & \xmark & \xmark \\
OEL~\citep{ye2026oel}
& \cmark & \cmark & \cmark & \xmark & \xmark \\
Inference-time memory agents
& \xmark & \xmark & \cmark & \cmark & \xmark \\
\midrule
\textbf{PMD}
& \cmark & \cmark & \cmark & \cmark & \cmark \\
\bottomrule
\end{tabular}
}
\vspace{-1.0em}
\end{wraptable}

Unlike work on reflection \cite{shinn2023reflexion}, refinement \cite{madaan2023selfrefine}, skill learning \cite{xia2026skillrl,wang2026skillsd,lu2026skill0}, and memory \cite{cheng2026mem2evolve}, which externalize feedback, trajectories, or skills and rely on them during inference, PMD uses memory only during teacher-path training. This design enables the student to internalize procedural knowledge, making memory a mechanism for learning rather than a dependency at deployment.

Our contributions are summarized as follows: 
\begin{enumerate}[leftmargin=*,topsep=0pt,itemsep=0pt]
    \item \textbf{Procedural Memory Distillation (PMD)}. We propose a self-distillation framework that turns episode-local supervision into cross-episode procedural memory, built online from the model's own rollouts and distilled into the policy's weights. The resulting model reasons natively at inference, with no external memory dependency.
    \item \textbf{Co-evolution as the design principle.} {\method} updates policy and memory jointly: rollouts from the current policy refresh memory, and the refreshed memory shapes the supervision that trains the next policy. This online coupling distinguishes PMD from static or offline memory banks.
    \item \textbf{A three-level procedural memory hierarchy.} We propose experience, insight, and behavior memory, and characterize the fidelity-transfer trade-off across them. Empirically, distilling Level-1 insights together with Level-2 behaviors yields the strongest internalized policy.
    \item \textbf{Empirical validation on two verifiable domains.} {\method} improves over GRPO and SDPO with both Qwen3-8B and OLMo3-Instruct-7B by $3.8$--$5.5\%$ on \textsc{SciKnowEval} and $7.9$--$13.6\%$ on \textsc{LiveCodeBench}. Freezing either the memory or the policy alone costs more than 10\% drop on \textsc{SciKnowEval}, isolating co-evolution as the source of PMD's gains.
    \item \textbf{Better test-time compute scaling.} We show that PMD continues to gain from additional rollouts where SDPO saturates, widening the gap with the rollout budget and opening 2-4$\times$ wider verifier headroom on \textsc{SciKnowEval}.
\end{enumerate}

%% file: related_work.tex
\section{Related Work}
\label{sec:related}

\paragraph{On-policy and self-distillation.}
Knowledge distillation trains a student to imitate a teacher distribution \citep{hinton2015distilling}, but language-model training also suffers from exposure bias when supervision is detached from the learner’s own states. This motivates online imitation and on-policy distillation, where supervision is generated on trajectories visited by the current policy \citep{ross2011dagger,agarwal2024opd,gu2024minillm,lu2025onpolicydistillation,song2026surveyopd}. Recent reasoning-oriented methods use privileged feedback, traces, or revisions for denser supervision: \sdpo{} learns from feedback-conditioned hindsight \citep{hubotter2026sdpo}, OPSD and OPSDC distill privileged reasoning context \citep{zhao2026opsd,sang2026opsdc}, SD-Zero converts binary rewards into self-revision targets \citep{he2026sdzero}, and RLSD studies self-distilled RLVR \citep{yang2026selfdistilledrlvr}. However, self-distillation can fail when teacher and student distributions drift, privileged information leaks, or reasoning verbalization is suppressed \citep{kim2026selfdistillationdegrade,li2026rethinkopd,shenfeld2026sdft,zhang2026ssdcode}. \method{} adopts the on-policy distillation view but changes the privileged signal: beyond the current answer, feedback, or trace, the teacher also receives procedural memory accumulated from prior attempts.

\paragraph{Experiential learning and internalizing context.}
Related work studies how models turn interaction history into reusable knowledge. \opcd{} distills transient context into model parameters via on-policy reverse-KL training \citep{ye2026opcd}, while \oel{} extracts experiential knowledge from deployment trajectories before consolidation \citep{ye2026oel}. ERL, X-KD, MR-Search, and R-Zero likewise use reflection, environments, or generated tasks to improve behavior across episodes \citep{shi2026erl,cai2026xkd,xiao2026mrsearch,huang2025rzero}. Our setting instead resembles training-time self-improvement: the policy revisits related problems across epochs, updates memory online, and distills from a memory-conditioned teacher into a memory-free student.

\paragraph{Memory, skills, and self-improvement.}
Memory-based agents maintain explicit stores of reflections, trajectories, or skills retrieved at inference time, including retrieval-augmented generation and long-term memory systems \citep{lewis2020rag,park2023generativeagents,packer2023memgpt}, reflection and skill-library agents \citep{yao2023react,shinn2023reflexion,madaan2023selfrefine,wang2023voyager}, and memory-management methods that learn to add, update, compress, or route memory \citep{yan2025memoryr1,yu2026agemem,wang2025memalpha,wu2025tokmem,zhang2026memfly}. Related work such as ReasoningBank, SkillRL, Skill-SD, SKILL0, MemGen, Mem2Evolve, EvolveR, and SkillOrchestra distills experience into reusable memories or skills for later reasoning or agent behavior \citep{ouyang2025reasoningbank,xia2026skillrl,wang2026skillsd,lu2026skill0,zhang2025memgen,cheng2026mem2evolve,wu2025evolver,wang2026skillorchestra}. Unlike these approaches, \method{} treats memory as a training scaffold rather than an inference-time dependency, studying which memory granularity can be distilled into the policy and transfer to held-out problems.

%% file: 2_method_v3.tex
\section{Method}
\label{sec:method}

\begin{figure*}[t]
    \centering
    \resizebox{\textwidth}{!}{\includegraphics{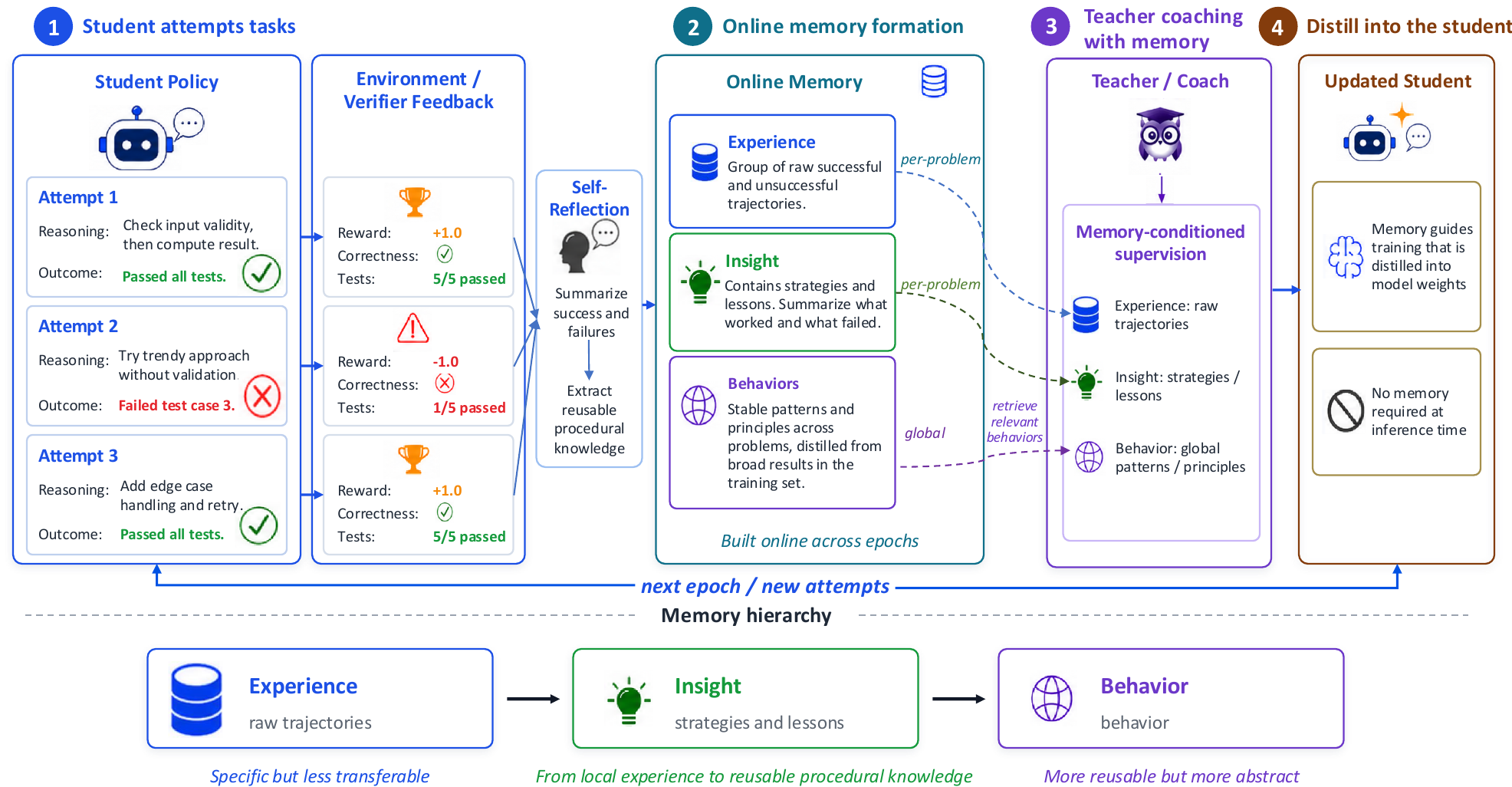}}
    \caption{\textbf{Overview of Procedural Memory Distillation (PMD).} (1) The student makes repeated attempts and receives verifier feedback. (2) Self-reflection summarizes successes and failures into online memory. (3) The teacher retrieves relevant memory in the form of experience, insight, and behaviors to provide memory-conditioned supervision. (4) This guidance is distilled into the updated student for the next epoch. The bottom row shows the memory hierarchy, from raw trajectories to increasingly abstract and reusable procedural knowledge.}
    \label{fig:pmd_overview}
\end{figure*}

Our goal is to turn repeated training-time experience into a form of procedural memory that improves the policy beyond the current episode. We build on SDPO and variants \citep{hubotter2026sdpo,zhao2026opsd}, which uses the current policy in two roles: a \emph{student} that produces the rollout and a \emph{self-teacher} that re-evaluates the same rollout after seeing additional training-time context such as environment feedback or a successful  solution. More broadly, this follows the on-policy distillation principle that the teacher should supervise the learner on states the learner actually visits, rather than providing off-policy demonstrations \citep{ross2011dagger,song2026surveyopd}. In standard SDPO, this extra context is largely episode-specific -- it is constructed from the current rollout group and is discarded after the current policy update. In this work, we take a broader perspective and hypothesize that such information across episodes captures information about the dynamics of how the model solves the same or similar problems over times, accumulating a rich source of self-learning signals.

We extend this setup by introducing \emph{online procedural memory distillation}. As shown in Figure~\ref{fig:pmd_overview}, this corresponds to a four-stage loop: (1) the student makes attempts to a problem  and receives verifier feedback; (2) self-reflection summarizes these successes and failures into different levels of procedural memory in the form of strategies, lessons, and behaviors; (3) the teacher leverages these problem-specific and global memory items to produce memory-conditioned supervision; and (4) this guidance is distilled back into the student, through self-distillation, for the next epoch. 

\subsection{Procedural Memory-conditioned Self-Distillation}
\label{sec:pmd_distillation}

We formalize PMD at the level of training update steps. Let $\pi_{\theta_t}$ denote the policy before update step $t$, and let
$M_t=\{M_t^{\mathrm{exp}},M_t^{\mathrm{ins}},M_t^{\mathrm{beh}}\}$ denote the procedural memory available at that step. Experience and insight memories are problem-specific, while behavior memory is shared across problems. We assume each epoch contains $K$ update steps.

Given a mini-batch $B_t$, the current policy produces rollout groups
\begin{equation}
\mathcal{T}_{i,t}
=
\left(
x_i,
\{y_{i,t}^{(j)}\}_{j=1}^{J},
\{r_{i,t}^{(j)},f_{i,t}^{(j)}\}_{j=1}^{J}
\right),
\qquad x_i\in B_t ,
\label{eq:rollout_group}
\end{equation}
where $r_{i,t}^{(j)}$ and $f_{i,t}^{(j)}$ denote the reward and feedback for rollout $j$. The same rollout group is used to update memory and to provide current-batch context for the teacher reprompt.

\paragraph{Memory update.}
Experience memory is updated online as new rollouts arrive. After the rollout group for problem $x_i$ is available, PMD updates the problem-specific experience and insight memories as
\begin{equation}
M_{t+1}^{\mathrm{exp}}[i]
=
\mathcal{U}^{\mathrm{exp}}
\left(
M_t^{\mathrm{exp}}[i],
\mathcal{T}_{i,t}
\right),
\qquad
M_{t+1}^{\mathrm{ins}}[i]
=
\mathcal{U}^{\mathrm{ins}}
\left(
M_t^{\mathrm{ins}}[i],
\mathrm{Reflect}(\mathcal{T}_{i,t})
\right),
\qquad x_i\in B_t .
\label{eq:problem_memory_update}
\end{equation}
For problems not visited in the current mini-batch, their experience and insight memories remain unchanged. The reflection step extracts strategies from successful rollouts and lessons from failed rollouts; when both are available, it contrasts successes and failures to identify what distinguishes correct reasoning from incorrect reasoning.

Behavior memory is updated on a slower time scale. Within an epoch, it is kept fixed. After every $K$ update steps, PMD clusters semantically related training questions, aggregates their latest insight memories, and abstracts them into reusable behavioral patterns:
\begin{equation}
M_{t+K}^{\mathrm{beh}}
=
\mathcal{U}^{\mathrm{beh}}
\left(
M_t^{\mathrm{beh}},
\left\{
\mathrm{Abstract}
\left(
\{M_{t+K}^{\mathrm{ins}}[i] : i \in c\}
\right)
:
c \in \mathrm{Cluster}(\mathcal{D}_{\mathrm{train}})
\right\}
\right).
\label{eq:behavior_memory_update}
\end{equation}
Thus, experience memory is updated as rollouts arrive, insight memory is updated after each problem-level rollout group, and behavior memory is consolidated periodically across related problems.

\paragraph{Memory-conditioned teacher.}
The teacher is conditioned on the latest available memory. For each problem $x_i$, PMD directly accesses its problem-specific memories and retrieves relevant behaviors from the global behavior bank:
\begin{equation}
m_{i,t}^{\mathrm{exp}}=M_{t+1}^{\mathrm{exp}}[i],
\qquad
m_{i,t}^{\mathrm{ins}}=M_{t+1}^{\mathrm{ins}}[i],
\qquad
m_{i,t}^{\mathrm{beh}}=\mathcal{R}(M_t^{\mathrm{beh}},x_i).
\label{eq:memory_access}
\end{equation}
The teacher memory context is then
\begin{equation}
m_{i,t}
=
\mathrm{Compose}
\left(
m_{i,t}^{\mathrm{exp}},
m_{i,t}^{\mathrm{ins}},
m_{i,t}^{\mathrm{beh}}
\right),
\label{eq:memory_compose}
\end{equation}
where $\mathcal{R}$ retrieves relevant behavior memories conditioned on the current problem.

Let $g_{i,t}$ denote the current-batch context from $\mathcal{T}_{i,t}$, such as verifier feedback or a successful sibling response. The memory-conditioned self-teacher is
\begin{equation}
q_{\theta_t}(\cdot \mid x_i,m_{i,t},g_{i,t})
:=
\pi_{\theta_t}
\left(
\cdot \mid \mathrm{reprompt}(x_i,m_{i,t},g_{i,t})
\right).
\label{eq:memory_teacher}
\end{equation}
As in SDPO, the teacher is fixed through $\mathrm{stopgrad}$. PMD differs by conditioning the teacher on procedural memory updated from the current policy's attempts, rather than relying only on episode-local feedback.

\paragraph{Policy update.}
Following SDPO, the student is trained on trajectories sampled from the current policy and matches the memory-conditioned teacher on the same prefix states:
\begin{equation}
\mathcal{L}_{\mathrm{PMD}}(\theta;\theta_t,M_t)
=
\mathbb{E}_{x_i,y_i}
\left[
\sum_{s=1}^{|y_i|}
\mathrm{KL}\!\left(
\pi_{\theta}(\cdot \mid x_i,y_{i,<s})
\;\middle\|\;
\mathrm{stopgrad}
\left(
q_{\theta_t}(\cdot \mid x_i,m_{i,t},g_{i,t},y_{i,<s})
\right)
\right)
\right],
\label{eq:pmd_loss}
\end{equation}
where $x_i\in B_t$ and $y_i$ is a rollout from $\mathcal{T}_{i,t}$. The policy is updated as
\begin{equation}
\theta_{t+1}
\leftarrow
\theta_t
-
\eta
\nabla_{\theta}
\mathcal{L}_{\mathrm{PMD}}(\theta;\theta_t,M_t).
\label{eq:policy_update}
\end{equation}

Feedback from the current rollout explains a local error or success, while procedural memory summarizes what the learner has discovered across repeated attempts: recurring mistakes, reusable strategies, and higher-level behaviors that may not be visible from a single batch. Because the policy changes during training, the memory that guides the teacher must also evolve. A memory bank written once by an earlier policy can drift out of sync with the learner's current policy. This motivates building memory online, so that the teacher remains compatible with the policy it supervises~\cite{li2026rethinkopd} and the knowledge source stays consistent with the on-policy learner that consumes it~\cite{ye2026oel}.

\subsection{Online Construction of Procedural Memory}
\label{sec:method:lifecycle}

PMD organizes procedural memory into three levels: \emph{experience}, \emph{insight}, and \emph{behavior} memory. During training, the model interacts with the environment through trial and error. Each rollout provides a training signal, such as reward, feedback, or a successful sibling response, and PMD records these signals as experience memory for each problem. The agent then reflects on the accumulated experience for the same problem to extract insights, including strategies that led to success and lessons that explain failure. This is similar in spirit to off-policy learning, where an agent can learn from past experience rather than only from the latest on-policy action \citep{watkins1992q}. Finally, PMD abstracts across insights from different problems to produce behavior memory: reusable skills or reasoning patterns that recur across problems. Thus, the memory hierarchy moves from concrete attempts, to problem-level insights, to cross-problem behaviors.

\textbf{Experience memory.}
As rollouts are produced by the current policy $\pi_{\theta_k}$, PMD stores raw experience on a per-problem basis. For each training example $x_i$, the experience memory $M_k^{\text{exp}}[i]$ records concrete attempts, including successful rollouts, failed rollouts, rewards, and environment feedback if available. This is the most faithful memory level: it preserves model's detailed reasoning. Because experience memory is updated online after each batch, it tracks the evolving policy rather than freezing the perspective of an earlier model.

\textbf{Insight memory.}
Experience memory is faithful but often too local and verbose to guide future learning directly. PMD therefore converts stored experience into problem-level insights. For each problem $x_i$, the insight memory $M_k^{\text{ins}}[i]$ summarizes successful and unsuccessful rollouts into strategies and lessons. Strategies capture reasoning patterns that consistently produce correct answers, while lessons identify recurring mistakes and explain their failures. When both successful and failed attempts are available, extraction is contrastive: the model compares the two groups to identify what differentiates successful reasoning from unsuccessful reasoning, rather than summarizing each rollout independently. PMD also uses the fraction of successful rollouts as a confidence signal, making insight memory reusable, confidence-aware training guidance across epochs.

\textbf{Behavior memory.}
The final level abstracts across problems. PMD encodes each question with an embedding model, clusters questions by similarity, and uses an LLM-based abstraction module to distill the experiences and insights within each cluster into behavior memory $M_k^{\text{beh}}$. Each behavior is a short reusable instruction that captures recurring reasoning patterns, mistakes, or skills across related questions, providing guidance beyond a single training example. This completes the memory hierarchy from experience to insight to behavior.

The hierarchy reflects a trade-off between concrete and abstract memories. Experience memory preserves faithful evidence but remains local; behavior memory transfers broadly but can become coarse; insight memory balances both by preserving compact lessons. PMD studies which memory level best supports learning and generalization. See Appendix Sections~\ref{app:insight-prompts}, \ref{app:behavior-prompts}, and \ref{app:memory-examples} for details.

\paragraph{Memory access.}
PMD accesses memory according to its scope. Experience and insight memories are problem-specific and are read for the current training example. Behavior memory is global and is retrieved by $\mathcal{R}$, a dense retrieval function that encodes the current question $x_i$ with an embedding model~\cite{zhang2025qwen3} and returns the top-$K$ most similar behavior entries from the global bank. This keeps lower-level memory grounded in the current problem while allowing behavior memory to transfer reusable procedural guidance across related problems. Memory access occurs only on the teacher path. Rather than creating a memory-augmented inference model, PMD uses memory to let the teacher convert accumulated experience into a stronger target distribution for the student. This distinguishes PMD from retrieval-based memory systems such as RAG~\citep{lewis2020rag}, MemGPT~\citep{packer2023memgpt}, and Memento~\citep{zhou2025memento}, which rely on memory at inference time. In PMD, the student requires no inference-time memory, as procedural knowledge is gradually internalized through self-distillation.

%% file: 3_experiment.tex
\section{Experiments}
\label{sec:exp}

\subsection{Experimental setup}
\paragraph{Data and Metrics} We evaluate \method{} in two verifiable domains: \textbf{\textsc{SciKnowEval}}, a science multiple-choice reasoning benchmark\citep{feng2024sciknoweval} covering biology, chemistry, physics, and materials science, and \textbf{LiveCodeBench}, a contamination-aware code-generation benchmark with execution-based unit-test feedback \citep{jain2024livecodebench}. Following SDPO \cite{hubotter2026sdpo}, we report \textbf{avg@16} on \textsc{SciKnowEval}, where each rollout is scored as $1.0$ if the extracted answer letter matches the ground truth and $0.0$ otherwise, and \textbf{score@4} on LiveCodeBench, where sparse reward assigns $1.0$ only when all unit tests pass. Both settings produce repeated successes, failures, and feedback across epochs, allowing us to study whether episode experience can be compressed into memory and distilled into the policy.

\paragraph{Model and Training} We evaluate \method{} with two open-source instruction-tuned policies, Qwen3-8B \citep{qwen2025qwen3} and OLMo3-Instruct-7B \citep{olmo2025olmo}. Unless otherwise stated, all experiments are conducted with thinking mode disabled (\textit{think-off}) following the setup in SDPO \citep{hubotter2026sdpo} for fair comparison. During training, we sample $n=8$ on-policy rollouts per prompt and use the same \sdpo{}-style reverse-KL self-distillation objective. Appendix~\ref{app:exp-details} provides additional details on data splits, infrastructure, hyperparameters, memory construction, and evaluation protocols.

\paragraph{Baselines}
We compare PMD against the base policy, GRPO~\citep{shao2024deepseekmath,guo2025deepseekr1}, and SDPO~\citep{hubotter2026sdpo}. GRPO isolates sparse reward-based optimization, while SDPO isolates feedback-conditioned self-distillation without explicit procedural memory. PMD uses the same self-distillation backbone as SDPO, but conditions the teacher on online procedural memory built from the evolving policy's own trajectories.

\subsection{Procedural memory provides strong self-learning signal}

\paragraph{Main results.}
In Table~\ref{tab:main_results}, we show that {\method} consistently outperforms both GRPO and SDPO across model families and benchmarks. Relative to SDPO, PMD improves \textsc{\textsc{SciKnowEval}} AVG from $74.4$ to $77.2$ for Qwen3-8B and from $69.5$ to $73.3$ for OLMo3-Instruct-7B, corresponding to relative gains of $3.8\%$ and $5.5\%$, respectively. On \textsc{LiveCodeBench}, PMD improves over SDPO from $47.9$ to $51.7$ for Qwen3-8B and from $45.0$ to $51.1$ for OLMo3-Instruct-7B, yielding relative gains of $7.9\%$ and $13.6\%$. Since the key difference between PMD and SDPO is whether model rollouts are accumulated as memory during training, these results suggest that training-time memories provide a valuable self-learning signal that SDPO does not exploit. In the rest of this section, we present analytical studies to better understand the source of these gains.

\begin{table}[t]
\centering
\caption{
\textbf{Main results on \textsc{SciKnowEval} and LiveCodeBench.} We compare PMD with the base policy, GRPO, and SDPO. All experiments use thinking-mode disabled (\textit{think-off}). \textsc{SciKnowEval} reports avg@16, where each rollout is scored as correct if the extracted answer letter matches the ground truth. LiveCodeBench reports score@4 with sparse unit-test rewards, where a rollout receives score $1.0$ only if all unit tests pass. \textsc{SciKnowEval} AVG is the mean over biology, chemistry, physics, and materials. Best results are shown in bold.
}
\label{tab:main_results}
\resizebox{\linewidth}{!}{
\begin{tabular}{llcccccc}
\toprule
\multirow{2}{*}{Model} & \multirow{2}{*}{Method} 
& \multicolumn{5}{c}{\textbf{SciKnowEval}} 
& \multicolumn{1}{c}{\textbf{LiveCodeBench}} \\
\cmidrule(lr){3-7} \cmidrule(lr){8-8}
& & Biology & Chemistry & Physics & Materials & AVG & v6 \\
\midrule
\multirow{4}{*}{Qwen3-8B}
& Base Policy~\citep{qwen2025qwen3} 
& 32.4 & 41.6 & 58.5 & 59.2 & 47.9 & 27.1 \\
& GRPO~\citep{shao2024deepseekmath,guo2025deepseekr1} 
& 63.2 & 73.2 & 70.6 & 70.7 & 69.4 & 41.2 \\
& SDPO~\citep{hubotter2026sdpo} 
& 63.6 & 80.6 & 72.8 & 80.4 & 74.4 & 47.9 \\
& \textbf{PMD} 
& \textbf{68.5} & \textbf{82.8} & \textbf{74.7} & \textbf{82.9} & \textbf{77.2} & \textbf{51.7} \\
\midrule
\multirow{4}{*}{OLMo3-Instruct-7B}
& Base Policy~\citep{olmo2025olmo} 
& 15.4 & 22.6 & 35.9 & 36.7 & 27.7 & 27.7 \\
& GRPO~\citep{shao2024deepseekmath,guo2025deepseekr1} 
& 45.8 & 70.1 & 63.3 & 76.5 & 63.9 & 36.1 \\
& SDPO~\citep{hubotter2026sdpo} 
& 54.1 & 79.0 & 66.1 & 76.8 & 69.5 & 45.0 \\
& \textbf{PMD} 
& \textbf{66.1} & \textbf{81.3} & \textbf{68.2} & \textbf{77.4} & \textbf{73.3} & \textbf{51.1} \\
\bottomrule
\vspace{-1em}
\end{tabular}
}
\end{table}

\begin{table}[t]
\addtolength{\tabcolsep}{-3pt}
\centering
\caption{
\textbf{Decomposing PMD's gain into interacting mechanisms.} %
\emph{PMD-Transient} is identical to \emph{PMD} but discards memory after every step, isolating within-step reflection from cross-step persistence. \emph{Evolving Memory + Frozen Policy} updates memory while keeping the policy fixed. \emph{Frozen Memory + SDPO} distills from a fixed memory bank, built once and held fixed. \textsc{SciKnowEval} reports avg@16 and LiveCodeBench reports score@4.
}
\label{tab:memory_ablation}
\resizebox{\linewidth}{!}{
\begin{tabular}{llcccccc}
\toprule
\multirow{2}{*}{Model} & \multirow{2}{*}{Method} 
& \multicolumn{5}{c}{\textbf{SciKnowEval}} 
& \multicolumn{1}{c}{\textbf{LiveCodeBench}} \\
\cmidrule(lr){3-7} \cmidrule(lr){8-8}
& & Biology & Chemistry & Physics & Materials & AVG & v6 \\
\midrule
\multirow{5}{*}{Qwen3-8B}
& Base Policy~\citep{qwen2025qwen3} 
& 32.4 & 41.6 & 58.5 & 59.2 & 47.9 & 27.1 \\

& SDPO~\citep{hubotter2026sdpo} 
& 63.6 & 80.6 & 72.8 & 80.4 & 74.4 & 47.9 \\
& Evolving Memory + Frozen Policy 
& 34.3 & 49.8 & 60.9 & 70.8 & 54.0 & 35.9 \\
& Frozen Memory + Evolving Policy 
& 44.6 & 77.7 & 63.1 & 74.5 & 65.0 & 47.5 \\
& PMD-Transient
& 67.4 & 81.8 & 73.8 & 79.8 & 75.7 & 48.1 \\
& \textbf{PMD} 
& \textbf{68.5} & \textbf{82.8} & \textbf{74.7} & \textbf{82.9} & \textbf{77.2} & \textbf{51.7} \\
\bottomrule
\vspace{-1em}
\end{tabular}
}
\end{table}

\paragraph{Decomposing the gain: reflection, persistence and co-evolution.}
{\method} improves learning through three interacting mechanisms: (\textit{i}) reflective abstraction from rollouts into reusable strategies and lessons, (\textit{ii}) persistence of these abstractions across optimization steps, and (\textit{iii}) co-evolution of memory and policy during training. To disentangle these effects, we introduce a set of controlled variants. \textbf{PMD-Transient} constructs memory from the current batch, used to condition the teacher for a single optimization step, and then discarded. It retains the full reflection pipeline---rollouts $\rightarrow$ experience $\rightarrow$ insights $\rightarrow$ behaviors $\rightarrow$ teacher conditioning---but removes cross-step accumulation. Comparing PMD-Transient with SDPO and {\method} isolates the contribution of reflective abstractions and persistent memory beyond transient reflection. To further study the role of co-evolution between memory and policy, we consider two additional variants. \textbf{Evolving Memory + Frozen Policy} keeps the base policy fixed while continuously updating the memory across epochs, testing whether memory accumulation alone can improve performance in the absence of parameter updates. In contrast, \textbf{Frozen Memory + Evolving Policy} first constructs a memory bank using the frozen-policy setup, and subsequently freezes this memory while training the policy using SDPO-style updates. This isolates whether a static repository of distilled strategies is sufficient, or whether gains arise specifically from the iterative co-adaptation between the evolving policy and evolving memory.

Table~\ref{tab:memory_ablation} decomposes {\method}'s improvement over SDPO. \textbf{Reflection helps even without persistence.} PMD-Transient discards memory after every step, yet still outperforms SDPO (+1.3pp on \textsc{SciKnowEval AVG} and +0.2pp on \textsc{LiveCodeBench}). Structured insights extracted from a single batch already provide a stronger signal than raw context. \textbf{Persistence drives most of the gain.} Allowing memory to persist across steps (PMD) yields an additional +1.5pp on \textsc{SciKnowEval} and +3.6pp on \textsc{LiveCodeBench}. The two mechanisms are complementary, but their relative importance depends on task structure: (1) in science MCQ, where a single batch often suffices to uncover the right strategy, reflection and persistence contribute roughly equally; (2) in code generation, where successful patterns are rarer and must be consolidated across many attempts, persistence accounts for nearly all the improvement. \textbf{Evolving memory alone is insufficient.} Updating memory while keeping the policy fixed barely improves performance (54.0 / 35.9): no distillation occurs, only a richer prompt that the model never internalizes. \textbf{Static memory loses alignment with an evolving policy.} Training against a memory bank constructed once at initialization performs substantially better (65.0 / 47.5), but still falls well short of PMD because the policy evolves while the memory remains fixed, causing the teacher signal to become stale. Together, these results identify co-evolution as the main driver of PMD's gains, with the value of persistence increasing alongside task difficulty.

\subsection{Transferability and Utility of Learned Memory}

\begin{wrapfigure}{r}{0.5\linewidth}
    \vspace{-1.5\baselineskip}   %
    \centering
    \includegraphics[
        width=\linewidth,
        trim={0 0 0 10},
        clip
    ]{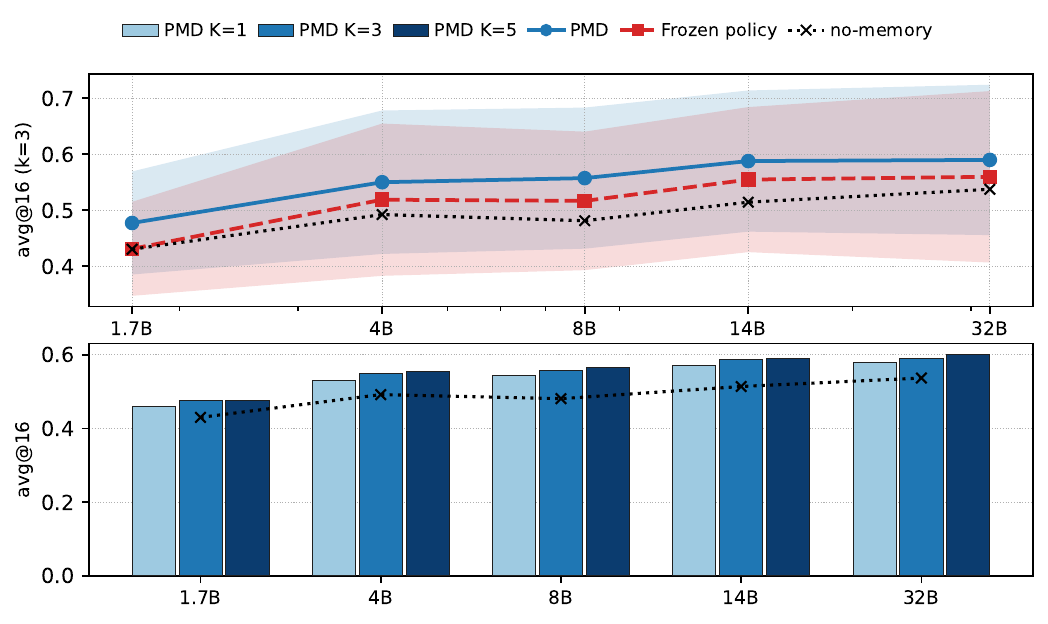}
    \caption{\textbf{Memory transfer on \textsc{SciKnowEval}} Memories are learned from Qwen3-8B under both PMD (co-evolving policy) and frozen-policy settings, then transferred across model scales. Top: PMD vs. frozen-policy memory transfer (shaded bands: cross-domain variability). Bottom: PMD retrieved memories vs.  performance ($K\in{1,3,5}$); dotted black denotes no-memory.}
    \label{fig:transferability}
        \vspace{-1em}
\end{wrapfigure}

We evaluate cross-scale transfer of learned memory (insight+behavior) on \textsc{SciKnowEval} to understand if it provides additive gains beyond policy quality. The memory is constructed from Qwen3-8B rollouts under both PMD (co-evolving memory) and frozen-policy settings, then evaluated against target models ranging from Qwen3-1.7B to Qwen3-32B. As shown in Figure \ref{fig:transferability}, across all sizes, memory-augmented inference outperforms the no-memory baseline, indicating that retrieved memory encodes reusable task signal rather than model-specific artifacts. Finally, PMD co-evolved memory outperforms the memory evolved with frozen policy.

Two patterns are robust. First, co-evolved memory (PMD) outperforms memory used with a frozen policy, showing that jointly adapting policy and memory improves downstream transfer quality. Second, increasing retrieval depth yields monotonic improvements (top-5 $>$ top-3 $>$ top-1), suggesting that additional retrieved memories are beneficial rather than noisy. Importantly, memory gains are large enough to offset model scale in multiple regimes: a smaller model with deeper retrieval can surpass a larger no-memory model (e.g., 4B with top-5 retrieval $>$ no-memory 8B), and this scaling trade-off continues at higher sizes (e.g., 8B with retrieval exceeding larger no-memory counterparts). Overall, these results establish that memory is transferable across scales and practically useful, and that co-evolving memory with policy is the strongest variant.

\subsection{Test-Time Scaling: PMD preserves the coverage that SDPO trades away}

\begin{figure}[t]
    \centering    
    \includegraphics[width=\linewidth]{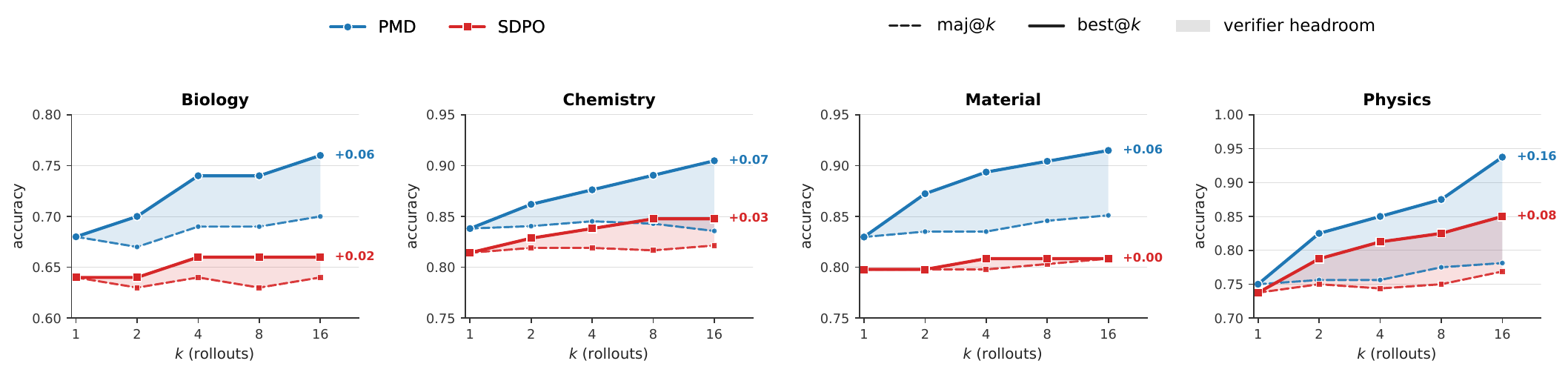}
    \caption{\textbf{PMD preserves answer-space coverage that SDPO collapses on \textsc{SciKnowEval}.} Using 16 rollouts/problem, lines show maj@k and best@k as rollout budget \(k\) increases. The shaded band (maj@k \(\rightarrow\) best@k) is verifier headroom. PMD's band is 2--4\(\times\) wider than SDPO across all subjects, indicating greater retained candidate diversity.}
    \label{fig:method_diversity_a}
    \vspace{-1em}
\end{figure}

\begin{figure}[t]
    \centering
    \includegraphics[width=\linewidth]{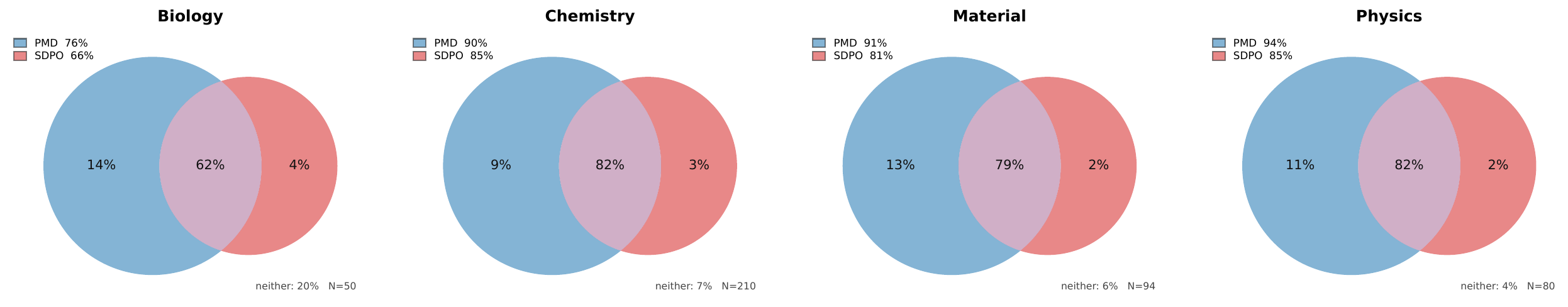}
    \caption{Per-subject Venn diagrams of problems with at least one of 16 rollouts correct ($N$ problems per panel; \emph{neither} = problems neither method solves). PMD's solved set covers 9--14\% more problems than SDPO's in every domain, while only 2--4\% are solved exclusively by SDPO. Together with verifier headroom, this shows that \textbf{memory-augmented training preserves coverage of the answer space, whereas SDPO collapses it onto a narrower subset}, limiting inference-time scaling.}
    \label{fig:method_diversity_b}
    \vspace{-1em}
\end{figure}

We sample $n{=}16$ rollouts per validation question and report \texttt{maj@k} (majority vote, random tie-break) and \texttt{best@k} (oracle pass-any). In Fig.~\ref{fig:method_diversity_a}, the shaded band between \texttt{maj@k} and \texttt{best@k} is \emph{verifier headroom}: the additional accuracy a perfect verifier could recover beyond self-consistency voting (annotated at $k{=}16$ as $+\Delta$). Across all \textsc{SciKnowEval} subjects, PMD preserves answer-space coverage that SDPO collapses. PMD is already better at $k{=}1$ by $2$--$5$ pp, and the gap grows to $7$--$10$ pp at $k{=}16$, because PMD continues improving with $k$ while SDPO saturates early. The \texttt{maj@k}$\rightarrow$\texttt{best@k} band is consistently wider for PMD (about $2$--$4\times$), indicating more recoverable accuracy for downstream reranking; on Material, SDPO’s headroom collapses to zero (\texttt{best@16}(=)\texttt{maj@16}), a direct signature of mode collapse. Consistently, Fig.\ref{fig:method_diversity_b} shows that PMD solves a strictly larger set of problems (coverage (+9)–(14\%) vs. SDPO, with only (2)–(4\%) SDPO-only solves), confirming that PMD expands useful answer-space coverage rather than merely shifting solved instances. Together, these results show that test-time scaling methods (self-consistency, verifier reranking, best-of-(N)) remain effective under PMD but are limited under SDPO collapse.

\subsection{Additional Analyses}
\label{sec:additional_analyses}
The appendix provides supporting analyses. Appendix~\ref{app:memory-level-ablation} compares experience-only memory, experience plus insight, and the full hierarchy with behavior memory, showing that procedural memory works best when distilled into the policy and that behavior memory gives the best average PMD performance. Appendix~\ref{app:memory-internalization} first probes memory internalization by decoding the student without memory and tracking the increased use of terms such as ``strategy'', ``lesson'', and ``behavior'' during training. Appendix~\ref{app:internalization} further analyzes internalization through response length and discourse structure, showing that PMD adopts teacher-side procedural reasoning patterns even when decoded without any memory prompt. Finally, Appendix~\ref{app:memory-dynamics} studies memory bank evolution, showing that lower-level memories collect problem-specific evidence while behavior memory acts as a higher-level consolidation mechanism.

%% file: 4_appendix_v3.tex
\clearpage
\section{More Experiment Analysis}
\subsection{Ablation on Memory Granularity}
\label{app:memory-level-ablation}

To study the effect of memory abstraction, we compare three memory configurations: \emph{experience only}, \emph{experience + insight}, and the full hierarchy of \emph{experience + insight + behavior}. This tests which level of procedural memory is most useful for reuse and internalization: concrete experience, problem-level insight, or cross-problem behavior \citep{xia2026skillrl,wang2026skillsd,lu2026skill0,ye2026oel}. We evaluate these configurations in two settings: \emph{Evolving Memory + Frozen Policy}, where memory can only help as external prompt context, and PMD, where memory is used to guide self-distillation and must be internalized into the policy to improve test performance.

\begin{table}[t]
\centering
\small
\caption{
Generalization of different procedural-memory levels on \textsc{SciKnowEval}. 
The frozen-policy setting tests whether memory helps as external context, while PMD tests whether each memory level can be distilled into the policy. 
AVG is the mean over the four subjects. Best result within each block is shown in bold.
}
\label{tab:memory_level_generalization}
\resizebox{\linewidth}{!}{
\begin{tabular}{llccccc}
\toprule
Setting & Memory Level 
& Biology & Chemistry & Physics & Materials & AVG \\
\midrule
\multirow{3}{*}{\begin{tabular}[c]{@{}l@{}}Evolving Memory\\+ Frozen Policy\\(Top-3)\end{tabular}}
& Experience Only 
& 42.3 & 53.5 & 59.1 & 68.8 & 55.9 \\
& Experience + Insight 
& \textbf{50.4} & 52.5 & \textbf{62.6} & 67.8 & \textbf{58.3} \\
& Experience + Insight + Behavior 
& 45.6 & \textbf{54.4} & 59.5 & \textbf{69.4} & 57.2 \\
\midrule
\multirow{3}{*}{PMD}
& Experience Only 
& 63.0 & 80.6 & 73.3 & 80.4 & 74.3 \\
& Experience + Insight 
& 67.1 & 82.1 & \textbf{75.2} & 78.2 & 75.7 \\
& Experience + Insight + Behavior 
& \textbf{68.5} & \textbf{82.7} & 74.1 & \textbf{83.6} & \textbf{77.2} \\
\bottomrule
\end{tabular}
}
\end{table}

Table~\ref{tab:memory_level_generalization} shows that procedural memory is most effective when it is distilled into the policy rather than used only as external context. In the frozen-policy setting, adding insight improves Biology and Physics, while adding behavior helps Chemistry and Materials; however, the gains are inconsistent because the underlying policy is unchanged. Under PMD, adding insight improves over experience-only memory on Biology, Chemistry, and Physics, suggesting that problem-level strategies and lessons provide a stronger training signal than concrete experience alone. Adding behavior gives the best average performance and achieves the strongest results on Chemistry and Materials, while remaining above experience-only memory on all four domains. These results support our fidelity--transfer trade-off hypothesis: experience preserves faithful local evidence, insight provides compact problem-level guidance, and behavior adds complementary cross-problem knowledge when distilled into the evolving policy.

\subsection{Internalization of Procedural Memory}
\label{app:memory-internalization}
\begin{figure}[t]
    \centering
    \includegraphics[width=\linewidth]{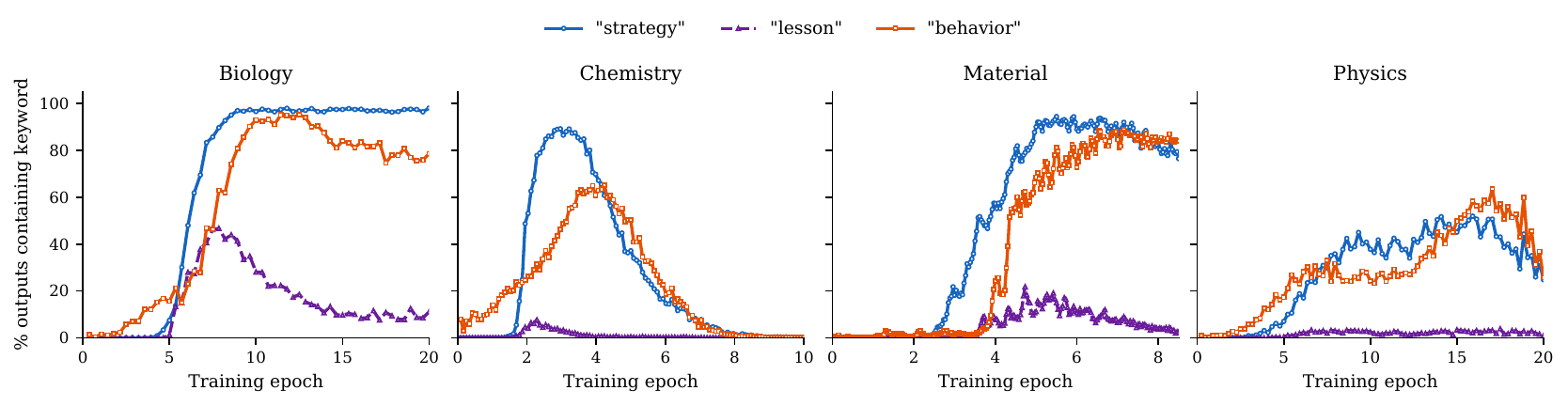}
    \caption{
    Surface-level probe of procedural-memory internalization. 
    We collect rollouts from student and track the percentage of rollouts containing keywords -- ``\texttt{strategy}'',``\texttt{lesson}'', and ``\texttt{behavior}'' -- across training steps. 
    The increasing frequency of memory-related terms, together with validation accuracy trends, suggests that PMD gradually transfers teacher-side procedural guidance into the student policy.
    }
    \label{fig:memory_internalization}
\end{figure}

We examine whether PMD can internalize procedural memory into the policy parameters. Prior work on self-distilled RLVR shows that privileged teacher-side information can be pathologically into model weights through distillation~\citep{yang2026selfdistilledrlvr}, making model hallucinate references to ground truth solution. PMD use this mechanism for a different purpose: the teacher is not only given the raw solution as feedback, but is also conditioned on abstract memory built from previous rollouts. 

At inference time, the student is decoded without any memory prompt. Therefore, if performance improves after training, the gain cannot come from reading external memory at test time. Instead, the useful information from procedural memory must have been absorbed into the model weights during distillation. Figure~\ref{fig:memory_internalization} provides a surface-level probe of this effect. As training proceeds, keyword used by different memory components -- ``\texttt{strategy}'',``\texttt{lesson}'', and ``\texttt{behavior}'' -- appear more often in the student rollouts, especially in Biology, Material, and Physics. The increase is biggest for ``\texttt{strategy}'' and ``\texttt{behavior}'', suggesting that the trained policy increasingly expresses the procedural concepts that were only available to the teacher during training. Together with improved validation performance, this supports our view of PMD as a training scaffold that converts abstract procedural guidance into persistent model behavior, rather than relying on solution leakage or inference-time memory access.

\begin{figure}[t]
    \centering

    \begin{minipage}{0.495\linewidth}
        \centering
        \includegraphics[width=\linewidth]{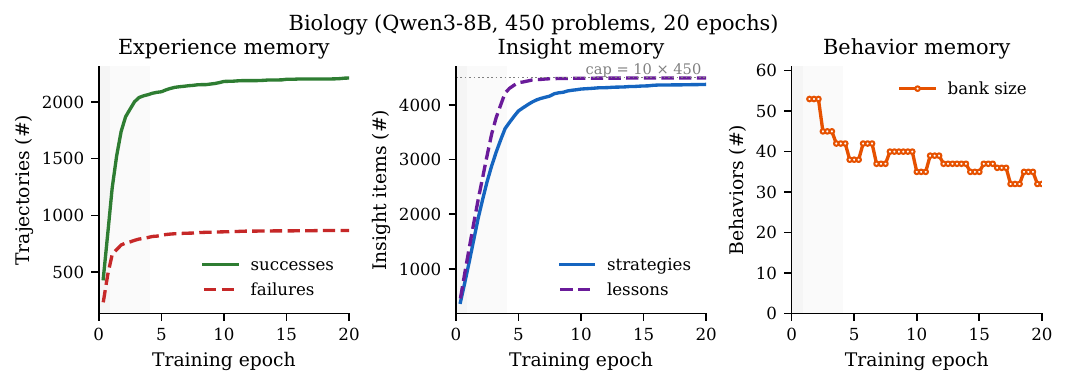}
    \end{minipage}
    \hfill
    \begin{minipage}{0.495\linewidth}
        \centering
        \includegraphics[width=\linewidth]{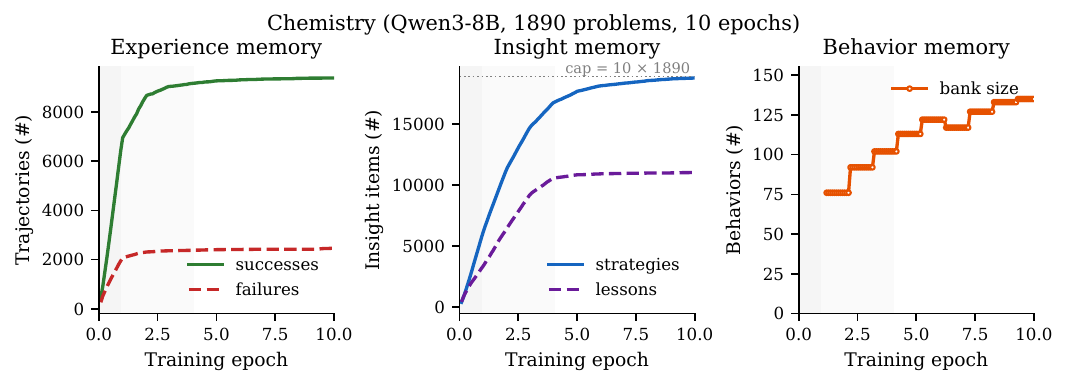}
    \end{minipage}
    \vspace{0.5em}
    \begin{minipage}{0.495\linewidth}
        \centering
        \includegraphics[width=\linewidth]{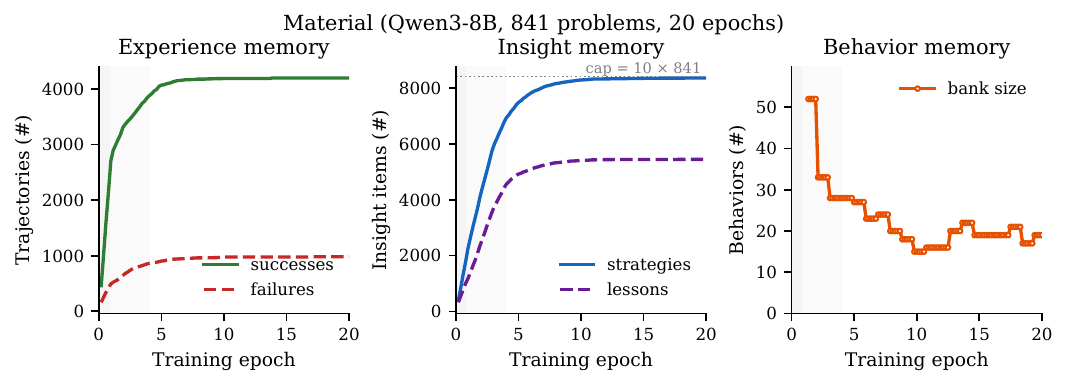}
    \end{minipage}
    \hfill
    \begin{minipage}{0.495\linewidth}
        \centering
        \includegraphics[width=\linewidth]{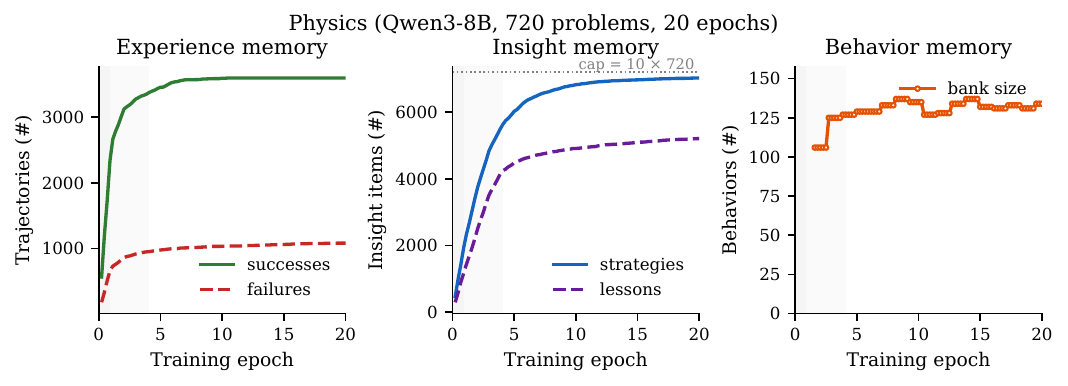}
    \end{minipage}

    \caption{
    Memory dynamics across \textsc{SciKnowEval} subjects. We track experience memory, insight memory, and behavior memory during PMD training. Per-problem memories tend to accumulate and saturate, while the global behavior bank shows subject-dependent consolidation dynamics.
    }
    \label{fig:memory-dynamics}
\end{figure}

\subsection{Length and discourse-level evidence of internalization.}~\label{app:internalization}
We further probe internalization by analyzing the length and discourse structure of decoded responses on \textsc{SciKnowEval}-Biology. At evaluation time, SDPO and PMD are decoded with the same no-memory prompt: neither model can retrieve strategies, lessons, or behaviors. Nevertheless, PMD produces moderately longer responses than SDPO while remaining far shorter than GRPO. As shown in Table~\ref{tab:length_analysis}, PMD improves accuracy from 63.6 to 68.5 over SDPO, while increasing median response length from 185 to 474 tokens. This increase is controlled rather than simply verbose: GRPO generates much longer responses, with a median length of 1306 tokens and generations often reaching the 8192-token cap, yet performs below PMD.

\begin{table}[t]
\centering
\caption{
Response length analysis on \textsc{SciKnowEval}-Biology. Accuracy follows the same evaluation
protocol as Table~\ref{tab:main_results}. Token statistics are computed over decoded evaluation
responses. PMD produces longer responses than SDPO, but remains well below the length regime of
GRPO.
}
\label{tab:length_analysis}
\resizebox{0.45\linewidth}{!}{
\begin{tabular}{lccccc}
\toprule
\textbf{Method} & \textbf{Acc.} & \textbf{Mean} & \textbf{Median} & \textbf{Max} & \textbf{Min} \\
\midrule
GRPO & 63.2 & 2457 & 1306 & 8192 & 492 \\
SDPO & 63.6 & 186 & 185 & 251 & 120 \\
PMD  & 68.5 & 481 & 474 & 711 & 296 \\
\bottomrule
\end{tabular}
}
\end{table}

To understand what accounts for the additional tokens, we manually annotate matched SDPO and PMD rollouts with the same inference prompt. Table~\ref{tab:discourse_internalization} shows that most of the extra PMD length comes from reusable reasoning machinery rather than generic elaboration. SDPO responses are dominated by a compact qualitative explanation and a final answer block. PMD responses add discourse patterns that mirror the training-time teacher context: per-option comparison and rejection, meta-level justification grounded in previously successful reasoning, and strategy-like phrasing that resembles behavior-memory entries. These components are not present in the inference-time prompt. They therefore provide qualitative evidence that PMD has absorbed teacher-side procedural structure into the policy weights.

\begin{table}[t]
\centering
\small
\caption{
Discourse-level decomposition of matched SDPO and PMD responses. Counts are approximate token contributions from a hand annotation. The additional PMD tokens mainly correspond to procedural reasoning patterns exposed only to the teacher during training.
}
\label{tab:discourse_internalization}
\resizebox{\linewidth}{!}{
\begin{tabular}{lccc}
\toprule
\textbf{Discourse function in response} 
& \textbf{SDPO} 
& \textbf{PMD} 
& \textbf{Likely source in training-time teacher} \\
\midrule
Qualitative reasoning over the concept or sequence
& 150 & 130 & Successful sibling rollout \\
Per-option comparison and rejection
& 0 & $\sim$120 & Contrastive strategies and lessons \\
Meta-level justification of the reasoning path
& 0 & $\sim$60 & Accumulated insight memory \\
Strategy phrasing resembling named behavior skills
& 0 & $\sim$80 & Behavior memory \\
Wrap-up paragraph restating the selected choice
& 0 & $\sim$40 & Successful-solution template \\
Final \texttt{<answer>} block
& 10 & 10 & Output format \\
\midrule
\textbf{Total}
& $\sim$160 & $\sim$440 & -- \\
\bottomrule
\end{tabular}
}
\end{table}

This analysis complements the keyword probe in Figure~\ref{fig:memory_internalization}. The keyword probe shows that PMD increasingly emits memory-related terms such as ``strategy'', ``lesson'', and ``behavior'' during training. The length and discourse analysis shows a more fine-grained version of the same phenomenon: the trained policy does not merely mention memory terms, but also adopts the procedural form of the teacher distribution. In this sense, PMD internalizes the memory bank as a reasoning style. The memory does not need to be available at test time because its reusable structure has been distilled into the student, at the cost of a moderate, sub-cap increase in response length.

\subsection{Memory Bank Size Dynamics.}
\label{app:memory-dynamics}
We track all three-level PMD memory during training on \textsc{SciKnowEval} 4 subjects with Qwen3-8B. As shown in Figure~\ref{fig:memory-dynamics}, across all \textsc{SciKnowEval} subjects, PMD exhibits a consistent separation between per-problem accumulation and cross-problem consolidation. Experience and insight memories grow rapidly during the early epochs as new problems are covered and repeated rollouts add successful trajectories, failed trajectories, strategies, and lessons. Their growth plateaus as the fixed per-problem capacity is approached, especially for lesson memory. In contrast, the behavior bank does not simply accumulate entries. Its dynamics depend on the diversity of the subject: biology and material quickly saturate and then shrink as redundant behaviors are removed, while chemistry and physics continue to admit new behaviors or maintain a larger bank. This suggests that lower-level memories mainly preserve problem-specific evidence, whereas the behavior bank acts as a higher-level consolidation mechanism that retains reusable skills only when they remain distinct and broadly useful.

\subsection{Compute Overhead}
\label{app:compute-overhead}
\begin{table}[t]
\centering
\small
\caption{
Training overhead relative to SDPO. We report median wall-clock time per training step and peak GPU memory under the same batch size ($32$), rollout count ($n=8$), and maximum response length ($8192$). Results are pooled across the four \textsc{SciKnowEval} subjects using Qwen3-8B on $8\times$H200 GPUs. PMD adds teacher-side memory retrieval and asynchronous memory extraction, but does not require additional teacher rollout generation.
}
\label{tab:compute_overhead}
\begin{tabular}{lccc}
\toprule
Method & Step Time & Relative Time & Peak GPU Memory \\
\midrule
SDPO & 37.0 s & $1.00\times$ & 54.6 GB \\
PMD  & 53.6 s & $1.45\times$ & 54.9 GB \\
\bottomrule
\end{tabular}
\end{table}
As shown in Table \ref{tab:compute_overhead}, PMD increases step time by $1.45\times$ while leaving peak GPU memory nearly unchanged $(+0.3$ GB$)$. The additional cost mainly comes from longer teacher prompts after memory injection, which increases generation and actor-update time. The memory bank itself is stored off-GPU as text and embeddings, and insight/behavior extraction runs asynchronously. Thus, PMD adds training-time overhead over SDPO, but preserves the same memory-free inference setting and does not introduce extra teacher rollout generation.

\section{Additional Method Details}
\label{app:method-details}

This appendix gives implementation details for the memory construction and teacher reprompting used by \method{}. The main text describes \method{} as an online lifecycle: the model records experience from its own attempts, reflects on that experience into insight, abstracts recurring patterns into behavior, uses the resulting memory to condition the teacher, and distills the teacher signal back into the policy. Here we make this lifecycle concrete and provide the prompt templates used by the memory modules.

\subsection{Memory store and update rules}
\label{app:memory-store}

The online memory is organized into three levels: \emph{experience}, \emph{insight}, and \emph{behavior}. These levels differ in scope and abstraction. Experience memory is the most concrete level and stores what the model actually tried on each problem. Insight memory reflects on the accumulated experience for the same problem and extracts compact problem-level guidance. Behavior memory abstracts across semantically related problems and stores reusable reasoning patterns in a global bank. During teacher reprompting, experience and insight are accessed directly through the current training example, while behavior is retrieved from the global behavior bank.

\paragraph{Experience memory.}
Each problem maintains a bounded set of successful and failed responses, together with rewards and environment feedback when available. In our current implementation, each memory entry stores up to five successful responses and up to three failed response--feedback pairs. Experience memory is updated synchronously after every training batch. Because naive storage quickly collapses into near-duplicate attempts, new responses pass through a diversity-aware novelty gate based on cosine similarity between response embeddings. Near-duplicates are rejected, and when storage is full, the most redundant stored item is evicted. This keeps experience memory faithful to the model's actual attempts while preserving a diverse set of evidence for later reflection.

\paragraph{Insight memory.}
After experience memory is updated, an asynchronous extractor reads the accumulated successes and failures for each problem and produces insight memory. Insight memory contains two types of problem-level items: \emph{strategies}, which describe reasoning patterns that led to correct solutions, and \emph{lessons}, which describe recurring mistakes and why they failed. When both successful and failed attempts are available, the extraction is contrastive: the extractor compares the two groups side by side and identifies what separates successful reasoning from unsuccessful reasoning. When only one side is available, the system falls back to a single-sided prompt. Insight items are accumulated across epochs rather than overwritten. New items are embedded, checked for novelty against existing insight items, and added only if they are sufficiently distinct. This allows the per-problem insight to become richer over time without collapsing into paraphrases.

\paragraph{Anti-shortcut filtering.}
For multiple-choice domains, the insight extractor includes explicit anti-shortcut rules. These rules forbid meta-language about successful or failed attempts, option-letter references, generic test-taking heuristics, and overly problem-specific phrasing. This matters in practice. In early biology runs on SciKnowEval, roughly $39\%$ of extracted insight items were shortcut-like. After adding the anti-shortcut rules to all extraction variants, the contamination rate dropped to roughly $3\%$. We measure contamination by scanning insight items with pattern-based filters for meta-language, option references, test-taking heuristics, and problem-specific phrasing.

\paragraph{Behavior memory.}
Behavior memory is a global bank of reusable instructions that transfer across problems. Each behavior contains a name, an instruction, its source, and an embedding used for retrieval. The bank stores up to 500 behaviors in the current implementation.

To construct behavior memory, PMD first encodes each question using Qwen3-Embedding-0.6B and clusters questions by semantic similarity in the embedding space. The LLM abstraction module is then applied to the experience and insight associated with each cluster. Thus, clustering is embedding-based rather than LLM-based; the LLM is used to distill each semantic cluster into compact behaviors. Each behavior captures a recurring reasoning pattern, mistake to avoid, or reusable skill that appears across related problems.

Behavior extraction runs after insight extraction and therefore builds primarily on problem-level insight rather than directly on noisy experience alone. During cold start, behaviors are extracted from semantic clusters built from the current available summaries. Once the behavior bank contains enough entries, we switch to a retrieve-then-decide update: the extractor first retrieves relevant existing behaviors, then decides whether to add new behaviors, revise existing ones, or remove outdated ones. This keeps the behavior bank curated and stable rather than allowing it to grow as an unstructured list of summaries.

\subsection{Teacher reprompt construction}
\label{app:teacher-reprompt}

PMD constructs the teacher reprompt by wrapping the original problem with the procedural memory accumulated so far. The memory context contains three types of information: problem-specific strategies extracted from successful rollouts, lessons extracted from failed rollouts, and retrieved behavior memory distilled from related problems. When available, the prompt also includes a previous successful solution and environment feedback from an unsuccessful attempt. The teacher is then asked to solve the original problem under this memory-conditioned context, and the student matches the resulting teacher distribution using the same reverse-KL self-distillation objective as in the main method.

\begin{promptbox}{Memory-Conditioned Teacher Reprompt}
\begin{PromptBlock}
{problem_text}

Relevant strategies from correct solutions to this problem:
{strategies}

Lessons from past failed attempts on this problem:
{lessons}

Reasoning skills --- reusable reasoning patterns learned across all
problems so far. Apply any that are relevant:

1. {behavior_1_name}: {behavior_1_instruction}
2. {behavior_2_name}: {behavior_2_instruction}
...
K. {behavior_K_name}: {behavior_K_instruction}

Correct solution:
{successful_previous_attempt}

The following is feedback from your unsuccessful earlier attempt:
{feedback_raw}

Correctly solve the original question.
\end{PromptBlock}
\end{promptbox}

Here, \texttt{\{problem\_text\}} is the original question. The fields \texttt{\{strategies\}} and \texttt{\{lessons\}} are problem-level insight memories extracted from accumulated successful and failed rollouts for the same problem. The behavior entries \texttt{\{behavior\_k\_name\}} and \texttt{\{behavior\_k\_instruction\}} are retrieved from the global behavior memory bank and represent reusable reasoning patterns learned across related problems.

The final two optional fields expose raw experience to the teacher. When a previous rollout solved the same problem, \texttt{\{successful\_previous\_attempt\}} provides that solution as a reference. When a failed rollout has executable or verifier-derived feedback, \texttt{\{feedback\_raw\}} provides the corresponding environment feedback. If either field is unavailable, the corresponding block is omitted from the reprompt.

The memory ablations differ only in which parts of this template are exposed to the teacher. In behavior mode, the teacher receives the original problem and retrieved behavior memory. In insight+behavior mode, it additionally receives the problem-specific strategies and lessons. In full mode, it receives the complete context, including strategies, lessons, retrieved behaviors, successful prior attempts, and available feedback. The memory extraction pipeline is unchanged across modes; only the visible memory context differs.

\subsection{Prompt templates for insight extraction}
\label{app:insight-prompts}

Insight extraction uses the student model itself, served separately, as a memory extractor. The prompts are designed to turn accumulated experience into reusable problem-level insight rather than problem-specific summaries. Each insight item is stored as either a strategy or a lesson.

\paragraph{Contrastive prompt.}
When both successful and failed attempts are available, we use a contrastive prompt. The prompt contains four ingredients: a confidence signal derived from rollout statistics, the original problem statement, a bounded set of successful attempts, and a bounded set of failed attempts with feedback. It then asks the extractor to produce strategies and lessons as the two fields of insight memory.

A simplified version of the prompt is shown below.

\begin{promptbox}{Contrastive Insight Extraction Prompt}
\begin{PromptBlock}
You are a science tutor extracting GENERAL scientific knowledge from a student model's attempts at a multiple-choice question. The goal is to produce reusable knowledge that helps solve ANY similar problem in this domain, not just this specific question.

Confidence signal: {n_success} out of {n_total} attempts were correct ({confidence_label}).

Question and Choices:
{problem_statement}

SUCCESSFUL ATTEMPTS -- {n_success}/{n_total} correct:
{successes_text}

FAILED ATTEMPTS -- {n_failure}/{n_total} wrong:
{failures_text}

Your task: extract GENERAL, TRANSFERABLE scientific principles revealed by comparing the successful and failed attempts. The output must be useful for solving other problems in the same domain, not just this specific question.

Rules:
- strategies: 2--3 general scientific principles or reasoning frameworks.
- lessons: 2--3 common misconceptions or reasoning errors.
- Be domain-specific.
- Each item: short title (<=10 words) + 2--4 sentences.
- Do not reference this question, option letters, attempts, the model, or test-taking shortcuts.

Respond ONLY with valid JSON:
{"strategies": [{"title": "...", "content": "..."}, ...],  "lessons": [{"title": "...", "content": "..."}, ...]}
\end{PromptBlock}
\end{promptbox}

The confidence signal calibrates how strongly the extractor should trust the observed pattern. The anti-shortcut rules prevent the extractor from producing answer-choice heuristics instead of reusable domain knowledge.

\paragraph{Single-sided fallbacks.}
When only successful attempts are available, the extractor produces strategy-style insight. When only failed attempts are available, it produces lesson-style insight. Both fallbacks preserve the same principle: outputs should be domain-specific, phrased as standalone general knowledge, and free of option references, attempt references, and shortcut heuristics.

\begin{promptbox}{Strategies-Only Insight Extraction Prompt}
\begin{PromptBlock}
Strategies-only fallback:

You are extracting GENERAL scientific knowledge from correct solutions to a multiple-choice question. The goal is to produce reusable principles that help solve ANY similar problem in this domain, not just this specific question.

Question and Choices:
{problem_statement}

Correct Solution(s):
{solutions}

Respond ONLY with valid JSON:
{"strategies": [{"title": "...", "content": "..."}, ...]}
\end{PromptBlock}
\end{promptbox}

\begin{promptbox}{Lessons-Only Insight Extraction Prompt}
\begin{PromptBlock}
Lessons-only fallback:

You are extracting GENERAL scientific misconceptions from wrong answers to a multiple-choice question. The goal is to identify common reasoning errors that recur across problems in this domain, not just errors specific to this question.

Question and Choices:
{problem_statement}

Wrong Solution(s) with Feedback:
{failures_text}

Respond ONLY with valid JSON:
{"lessons": [{"title": "...", "content": "..."}, ...]}
\end{PromptBlock}
\end{promptbox}

\paragraph{Shortcut contamination analysis.}
To quantify shortcut contamination, we scan extracted insight items with regex patterns corresponding to four categories: meta-language about successful or failed attempts, option-letter or option-value references, generic test-taking heuristics, and problem-specific phrasing such as ``this question'' or ``the given sequence.'' A memory item is flagged if it matches any of these categories, and contamination is reported as the fraction of flagged items in the store.

\subsection{Prompt templates for behavior extraction}
\label{app:behavior-prompts}

Behavior extraction operates across related problems rather than within a single problem. We first encode each question with Qwen3-Embedding-0.6B and group questions by semantic similarity. For each semantic cluster, we build a short summary consisting of the problem text and the available memory: primarily insight memory, with experience memory used as fallback evidence when insight is not yet available. 

\paragraph{Cold-start extraction.}
When the behavior bank is empty or still small, the extractor infers an initial set of reusable behaviors from the current semantic clusters.

\begin{promptbox}{Cold-Start Behavior Extraction Prompt}
\begin{PromptBlock}
You are a metacognitive strategist analyzing a cluster of semantically related problems and the model's accumulated memory on them. Your goal is to distill reusable behaviors -- general-purpose reasoning strategies, common pitfalls, and actionable rules -- that transfer across related problems.

A behavior is NOT a solution to a specific problem. It is a reusable skill, a pattern to recognize, a common mistake to avoid, or a strategy to apply. Behaviors should be general enough to help on future unseen problems.

Below are summaries from {n_problems} related problems:

{problem_summaries}

Instructions:
1. Identify recurring patterns across these related problems.
2. Extract 3--8 reusable behaviors, each with a name and instruction.
3. Focus on high-level, transferable guidance.
4. Include both positive behaviors and mistakes to avoid.
5. Avoid problem-specific wording, option-letter shortcuts, and references to individual attempts.

Respond ONLY with valid JSON:
{"behaviors": [{"name": "behavior_...", "instruction": "..."}, ...]}
\end{PromptBlock}
\end{promptbox}

\paragraph{Evolution prompt.}
Once the bank contains existing behaviors, the extractor switches to an evolution prompt. It is shown the current semantic-cluster summaries and the most relevant existing behaviors retrieved by embedding similarity. It then decides whether to create, update, or remove behavior entries.

\begin{promptbox}{Behavior Bank Evolution Prompt}
\begin{PromptBlock}
You are a metacognitive strategist maintaining a behavior bank -- a collection of reusable reasoning patterns, common pitfalls, and actionable rules that transfer across problems.

Below are summaries from {n_problems} semantically related problems:

{problem_summaries}

Here are EXISTING behaviors from your bank that are relevant to these problems:

{existing_behaviors}

Your task is to EVOLVE the behavior bank based on the new evidence:
1. Review existing behaviors against the new cluster evidence.
2. Identify gaps not covered by existing behaviors.
3. Decide on actions: new, update, or remove.
4. Keep behaviors reusable, concise, and independent of any single problem.

Respond ONLY with valid JSON:
{"actions": [
{"action": "new", "name": "behavior_...", "instruction": "..."},
{"action": "update", "name": "behavior_existing_name", "instruction": "..."},
{"action": "remove", "name": "behavior_bad_one"}]}
\end{PromptBlock}
\end{promptbox}

This retrieve-then-decide pattern prevents the behavior bank from becoming an uncurated list of summaries. Instead, the bank is maintained as a compact set of reusable behaviors that remain useful across training.

\subsection{Behavior retrieval and persistence}
\label{app:retrieval-persistence}

PMD accesses each memory level according to its scope. Experience and insight memories are problem-specific, so during training they are read directly for the current example. Behavior memory is cross-problem, so it is retrieved from the global behavior bank.

For behavior retrieval, we encode the current question with Qwen3-Embedding-0.6B and compare it with the embeddings of stored behaviors using cosine similarity. The top-$K$ most similar behaviors are returned and inserted into the teacher prompt. When feedback is available, we optionally append it to the retrieval query so that the retrieved behaviors can reflect both the problem and the current failure mode. In the main experiments, we retrieve the top-$K=3$ behavior entries.

All memory levels are periodically persisted to disk, including the experience store, insight store, global behavior bank, and validation outputs. The latest behavior bank is saved after each extraction step and automatically reloaded on restart. The final student is still evaluated without memory retrieval; behavior retrieval is used only on the teacher path during training.

\subsection{Qualitative Examples of Procedural Memory}
\label{app:memory-examples}

To make the memory hierarchy concrete, we show one example from \textsc{SciKnowEval} and one from \textsc{LiveCodeBench}. In both cases, experience preserves concrete attempts and feedback, insight compresses those attempts into problem-level strategies and lessons, and behavior abstracts recurring patterns across related problems.

\paragraph{SciKnowEval example.}
Table~\ref{tab:sciknow_memory_example} shows a chemistry question asking for the molar weight of the SMILES string \texttt{C1CC1C(=O)NCCC=O}. The correct answer is B, corresponding to a molar weight of 141.17 g/mol. The example illustrates how a concrete atom-counting failure becomes a reusable SMILES-parsing behavior.

\begin{table*}[t]
\centering
\small
\caption{
A \textsc{SciKnowEval} chemistry example showing the three PMD memory levels. Experience is problem-specific evidence; insight is a compact problem-level abstraction; behavior is a cross-problem rule distilled from a cluster of related SMILES molar-mass questions.
}
\label{tab:sciknow_memory_example}
\begin{tabular}{p{0.16\textwidth}p{0.78\textwidth}}
\toprule
Memory Level & Example \\
\midrule
Experience &
\textbf{Failure:} a rollout misreads the ring token \texttt{C1CC1}, overcounts carbon atoms, and selects A. The feedback records that the chosen answer is A while the correct answer is B.
\newline
\textbf{Success:} another rollout recognizes \texttt{C1CC1} as cyclopropane, accounts for the carbonyl, amide, and aldehyde groups, and selects B. \\
\midrule
Insight &
\textbf{Strategies:} count all atoms in the molecular formula, including implicit hydrogens omitted by SMILES; use precise atomic weights when answer options are close; decompose complex structures into functional groups before counting.
\newline
\textbf{Lessons:} avoid overcounting or undercounting atoms in rings; do not ignore implicit hydrogens in SMILES; verify ring tokens such as \texttt{C1CC1} rather than treating them as ordinary linear fragments. \\
\midrule
Behavior &
The behavior evolver clusters this problem with related SMILES molar-mass questions and produces reusable rules such as:
\newline
\textbf{\texttt{behavior\_account\_for\_implicit\_hydrogens}:} hydrogen atoms are often implicit in SMILES, especially in rings and functional groups, and must still be included in molar-mass calculations.
\newline
\textbf{\texttt{behavior\_verify\_ring\_atom\_counts}:} cyclic structures should be checked explicitly to avoid miscounting ring atoms. \\
\bottomrule
\end{tabular}
\end{table*}

\paragraph{LiveCodeBench example.}
Table~\ref{tab:lcb_memory_example} shows a code-generation problem asking for the capped geometric sum $X=\sum_{i=0}^{M} N^i$, printing $X$ if $X\leq 10^9$ and \texttt{inf} otherwise. The main traps are the degenerate case $N=1$, early termination that changes the sum, and modular exponentiation that destroys the exact value needed for a threshold comparison.

\begin{table*}[t]
\centering
\small
\caption{
A \textsc{LiveCodeBench} example showing the three PMD memory levels. Experience records concrete code failures and unit-test feedback; insight summarizes the local algorithmic mistakes; behavior abstracts them into reusable coding principles for capped arithmetic and closed-form formulas.
}
\label{tab:lcb_memory_example}
\begin{tabular}{p{0.16\textwidth}p{0.78\textwidth}}
\toprule
Memory Level & Example \\
\midrule
Experience &
The rollout group contains five successful and three failed attempts. One failed attempt uses an iterative loop with an off-by-one update and breaks early; the feedback includes cases such as $(N=21,M=5)$ where the output is too small. Another failed attempt uses modular exponentiation, e.g., \texttt{pow(N, M+1, 1e9+1)}, which changes the value and fails unit tests. A successful attempt handles $N=1$ separately, computes the closed form exactly without a modulus, and prints \texttt{inf} only after comparing the exact result to $10^9$. \\
\midrule
Insight &
\textbf{Strategies:} use the geometric-series formula when applicable; handle $N=1$ as a special case because the formula divides by $N-1$; use exact arithmetic for the final threshold comparison.
\newline
\textbf{Lessons:} modular arithmetic should not be used unless the problem asks for a value modulo some number; early termination is unsafe unless the maintained state proves the final capped comparison is unchanged; off-by-one additions can double-count the $N^0$ term. \\
\midrule
Behavior &
The behavior evolver clusters this problem with other capped-sum and closed-form arithmetic problems and produces reusable behaviors such as:
\newline
\textbf{\texttt{behavior\_handle\_formula\_degenerate\_cases\_separately}:} branch on algebraic degeneracies before applying a closed form, and reject modular arithmetic for overflow or cap decisions.
\newline
\textbf{\texttt{behavior\_cap\_growth\_during\_iterative\_arithmetic}:} when a numeric process is capped at a threshold, maintain the running value and stop only when the cap comparison is already determined. \\
\bottomrule
\end{tabular}
\end{table*}

These examples show the fidelity--transfer trade-off studied in the main paper. Experience is maximally faithful but tied to a single problem. Insight remains problem-specific but compresses repeated attempts into named strategies and lessons. Behavior removes most problem-specific details and can be retrieved for semantically related unseen problems, but it is necessarily coarser than the underlying experience.

\subsection{Algorithm}
\label{sec:method:algorithm}

Algorithm~\ref{alg:pmd} summarizes the PMD training loop.

\begin{algorithm}[t]
\caption{Procedural Memory Distillation}
\label{alg:pmd}
\begin{algorithmic}[1]
\State \textbf{Input:} dataset $\mathcal{D}$, student policy $\pi_\theta$, memory store $\mathcal{M}=\{M^{\mathrm{exp}}, M^{\mathrm{ins}}, M^{\mathrm{beh}}\}$
\For{epoch $=1,\dots,E$}
    \For{batch $B \subset \mathcal{D}$}
        \State Sample rollout groups $\{y_i\}_{i \in B}$ from $\pi_\theta(\cdot \mid x_i)$
        \State Obtain rewards and environment feedback
        \For{each problem $i \in B$}
            \State Update experience memory $M^{\mathrm{exp}}[i]$ with new successes, failures, rewards, and feedback
            \State Update insight memory $M^{\mathrm{ins}}[i]$ by reflecting on accumulated experience
        \EndFor
        \State Encode questions with embedding model and cluster related problems by semantic similarity
        \State Update behavior memory $M^{\mathrm{beh}}$ by abstracting cluster-level experience and insight
        \For{each problem $i \in B$}
            \State Access $m^{\mathrm{exp}}_i = M^{\mathrm{exp}}[i]$ and $m^{\mathrm{ins}}_i = M^{\mathrm{ins}}[i]$
            \State Retrieve $m^{\mathrm{beh}}_i = \mathcal{R}(M^{\mathrm{beh}}, x_i)$ using embedding model
            \State Compose teacher memory $m_i = \mathrm{Compose}(m^{\mathrm{exp}}_i, m^{\mathrm{ins}}_i, m^{\mathrm{beh}}_i)$
            \State Construct teacher context $\mathrm{reprompt}(x_i, f_i, m_i)$
        \EndFor
        \State Compute $\mathcal{L}_{\textsc{pmd}}$ and update $\theta$
    \EndFor
\EndFor
\State \textbf{Return:} distilled policy $\pi_\theta$
\end{algorithmic}
\end{algorithm}

The algorithm highlights three design choices. First, memory is constructed online from the model's own attempts under a changing policy, keeping it aligned with the learner. Second, memory is organized as a hierarchy of experience, insight, and behavior, allowing us to study the trade-off between fidelity and transfer. Third, memory is used only through the teacher path. The student does not rely on memory at inference time; retrieval strengthens the training target rather than creating a permanent serving dependency.

\section{Experimental Details}
\label{app:exp-details}

This section gives the implementation and evaluation details behind the experiments in Section~\ref{sec:exp}. We keep these details in the appendix so that the main experimental section can focus on the comparison among standard policy optimization, self-distillation, memory-only adaptation, fixed-memory distillation, and the full online \method{} training loop.

\subsection{Backbone and infrastructure}
\label{app:backbone-infra}

We train and evaluate two open-source instruction-tuned policies: Qwen3-8B \citep{qwen2025qwen3} and OLMo3-Instruct-7B \citep{olmo2025olmo}. Unless otherwise stated, both model families use the same PMD training recipe, reward functions, memory construction pipeline, and evaluation protocol. All runs use a single node with $8\times$H200-143GB GPUs. The policy is FSDP-sharded across all eight GPUs, and rollouts are served with SGLang v0.4 using the same shard. We colocate a second model server on GPU~7 in bf16 mode without CUDA graph to serve the student-contrastive memory extractor; this extractor uses the same backbone family as the policy being trained and shares the rollout/FSDP GPU through \texttt{mem-fraction-static=0.2}. Optimization uses AdamW in fp32, and rollout serving uses \texttt{gpu\_memory\_utilization=0.5}. For experiments with an evolving global behavior bank, the behavior-bank evolver is GPT-5.4 and is called at most once per training step after the warm-up period. Behavior clustering and behavior retrieval use Qwen3-Embedding-0.6B for both policy families.

\subsection{Datasets and rewards}
\label{app:datasets-rewards}

Table~\ref{tab:dataset_summary} summarizes the datasets. SciKnowEval uses held-out science multiple-choice questions and a rule-based reward that extracts the final answer letter \citep{feng2024sciknoweval}. LiveCodeBench uses half unit test of LiveCodeBench v6 is used for training, and the full v6 unit test set is reserved for evaluation \citep{hubotter2026sdpo}. The code reward scorer executes generated solutions against problem-specific unit tests and uses sparse rewards.

\begin{table}[t]
\centering
\small
\caption{Dataset splits and reward used in our experiments.}
\label{tab:dataset_summary}
\resizebox{\linewidth}{!}{
\begin{tabular}{llrrl}
\toprule
Benchmark & Domain & \#Train & \#Test & Reward \\
\midrule
SciKnowEval---Biology & MCQ & 450 & 50 & Rule-based final choice A--D \\
SciKnowEval---Chemistry & MCQ & 1,890 & 210 & Rule-based final choice A--D \\
SciKnowEval---Materials & MCQ & 841 & 94 & Rule-based final choice A--D \\
SciKnowEval---Physics & MCQ & 720 & 80 & Rule-based final choice A--D \\
LiveCodeBench v6 train / test & Code & 924 & 131 & Per-unit-test pass rate \\
\bottomrule
\end{tabular}
}
\end{table}

\subsection{Training hyperparameters}
\label{app:training-hparams}

We use the \sdpo{} objective with reverse KL between the rollout-conditioned student and a teacher reprompt enriched with retrieved memory \citep{hubotter2026sdpo}. Table~\ref{tab:training_hparams} lists the main hyperparameters. Entries that differ between SciKnowEval and LiveCodeBench are shown separately.

\begin{table}[t]
\centering
\small
\caption{Training hyperparameters for SciKnowEval and LiveCodeBench.}
\label{tab:training_hparams}
\resizebox{\linewidth}{!}{
\begin{tabular}{lll}
\toprule
Setting & SciKnowEval & LiveCodeBench \\
\midrule
Backbone & Qwen3-8B / OLMo3-Instruct-7B & Qwen3-8B / OLMo3-Instruct-7B \\
Total epochs & 20 & 20 \\
Optimizer & AdamW (fp32) & AdamW (fp32) \\
Learning rate & $1\times10^{-5}$ & $1\times10^{-6}$ \\
LR warmup steps & 10 & 0 \\
Train batch size & 32 & 32 \\
PPO mini-batch size & 32 & 1  \\
Rollouts per prompt $n$ & 8 & 8 \\
max prompt length & 2,048 & 2,048 \\
max response length & 8,192 & 8,192 \\
max model len & 18,944 & 18,944 \\
Rollout sampling & $T=1.0$, top-$p=1$, top-$k=-1$ & $T=1.0$, top-$p=1$, top-$k=-1$ \\
Validation samples & 16 samples & 4 samples \\
Validation interval & Every 5 steps & Every 5 steps \\
\sdpo{} $\alpha$ & 0.5 & 1.0 \\
distillation topk & 100 & 20 \\
EMA update rate & -- & 0.01 \\
Advantage estimator & GRPO, no critic & GRPO, no critic \\
Rollout-IS correction & Per-token & Per-token \\
Reasoning template & enable\_thinking=False & enable\_thinking=False \\
Reward & Rule-based MCQ & Sparse per-unit-test reward \\
\bottomrule
\end{tabular}
}
\end{table}

\subsection{Memory module and bank operating modes}
\label{app:memory-module-exp}

The teacher reprompt receives three memory levels in order: experience, insight, and behavior. Experience memory contains concrete student attempts and is always included in the full-memory setting, with each problem capped at five successful rollouts and three failed rollouts. Insight memory contains problem-level strategies and lessons distilled by the student-contrastive extractor; we use the MCQ extractor for SciKnowEval and the code extractor for LiveCodeBench. For each policy family, the extractor uses the same backbone family as the policy being trained. Extraction runs concurrently with \texttt{llm\_concurrency=4}, uses the \texttt{combined} conditioning mode, and accesses problem-keyed experience and insight memory exactly when available.

Behavior memory is a global bank of cross-problem behaviors. It is retrieved by dense similarity with Qwen3-Embedding-0.6B over the current problem text for both Qwen3-8B and OLMo3-Instruct-7B experiments. Because semantically similar tasks may still require different procedures, we use behavior retrieval only on the teacher path and keep the final student memory-free at inference time, partly to avoid the persistent imitation risks observed in experience-retrieval systems \citep{srivastava2025memorygraft}. We retrieve the top-$K=3$ behaviors per problem and inject them as a behavior-memory block.